# View-Based Luminance Mapping in Open Workplace

1ST GUANZHOU JI[1], 2ND TINGSONG OU[1], 3RD AZADEH O. SAWYER[1]
[1]Carnegie Mellon University, Pittsburgh, USA
[1]gji@andrew.cmu.edu, [2]tingsono@andrew.cmu.edu, [3]asawyer@andrew.cmu.edu

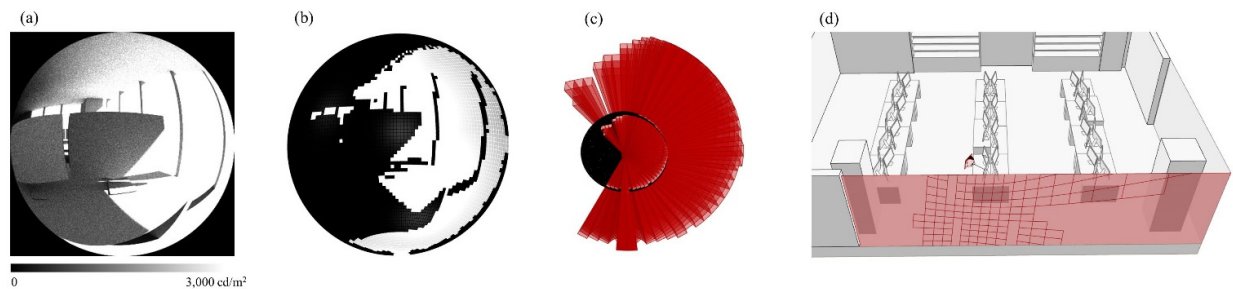

Fig. 1. Rendering and Luminance Mapping: (a) fisheye rendering in 180-degree Field of View (FOV); (b) pixel mapping from the image to spherical coordinates; (c) projection of high luminance values from the hemisphere onto the building facade; (d) facade areas outlined by the high luminance values from one view position.

## PART 1: PROPOSITION

### ABSTRACT

This paper introduces a novel computational method for mapping indoor luminance values on the facade of an open workplace to improve its daylight performance. 180-degree fisheye renderings from different indoor locations, view positions, and times of the year are created. These renderings are then transformed from two-dimensional (2D) images into three-dimensional (3D) hemispheres. High luminance values are filtered and projected from the hemisphere to the facade surface. This framework will highlight the areas of the facade that allow too much light penetration into the interior environment. The flexible workflow allows occupant-centric lighting analysis that computes multiple design parameters and synthesizes results for localized facade optimization and daylight design.

### 1. INTRODUCTION

The design of building facades significantly impacts building energy performance and human comfort. Although maximizing daylight in buildings is desirable, if not done correctly, it can lead to overheating of the space and occupants' visual discomfort and dissatisfaction. Several studies have investigated the influence of facade design on daylight, user satisfaction, and visual perception. These studies include light measurements and their impact on user preferences [Omidfar et al., 2015], daylight-driven interest using rendered scenes in Virtual Reality (VR) [Rockcastle et al., 2017], view accessibility [Turan et al., 2019], and user-defined visual preferences [Li and Samuelson, 2020]. A successful facade design ensures adequate light levels for indoor visual tasks and minimizes potential glare problems. As luminance is a key metric related to how the human eyes perceive brightness, luminance measurement becomes the critical variable in space design. Currently, computational simulation and field measurement are the two primary approaches for luminance analysis that can provide design suggestions for space layouts and window design. However, the results of conventional luminance analysis lack a three-dimensional representation that can be used to accurately trace and adjust the overlit areas. The concept of this study is to combine luminance simulation with view analysis and spatial projection to guide the facade design process.

### 2. LITERATURE REVIEW

#### 2.1. Luminance Study and Field Measurement

To evaluate light distribution in the real world, high dynamic range (HDR) photography is widely used in field measurement to capture the wide range of pixel values. Previous luminance studies used HDR images to explore diverse topics such as user's visual perception under direct sunlight [Jain et al., 2022], the impacts of different interior design elements (window height, seating position and location) on visual comfort [Kong et al., 2018], and development of glare metrics including Daylight Glare Probability (DGP) [Wienold and Christoffersen, 2006] and Unified Glare Probability (UGP) [Hirning et al., 2017]. When measuring the luminance distribution in the open workplace [Alicia and Simon, 2017] and private office room [Wymelenberg and Inanici, 2017], researchers computed glare metrics for different scenes and investigated the correlation between light levels and occupants' subjective responses. The glare results describe the probability of glare occurrence; however, field measurements are limited by the number of captured HDR images and respondents.

Visual comfort studies typically focus on the overlit areas within the occupants' field of view (FOV). When Van Den Wymelenberg et al. studied luminance distribution and occupant preferences, each captured scene was divided into several areas based on luminance threshold, solid view angle, and task surface [Van Den Wymelenberg et al., 2010]. Another study decomposed the fisheye imagery into a 16 by 16 grid matrix, allowing users to label the locations of glare sources in the surveys [Hirning et al., 2014]. These two studies divided the entire fisheye image into small areas for further luminance analysis. However, the results were based on users' annotation on the images rather than filtering pixel values computationally. In addition, a 180-fisheye perspective contains geometric distortion that results in deformation of the objects in the captured scene and will lose graphic fidelity near the edge region when image distortion is removed.

### 2.2. Computational Simulation

Computational approach allows batch simulation with high flexibility in view setting and time selection. Several experiments displayed 180-degree FOV images in VR headsets to study subjective visual perception in daylit spaces [Chamilothori et al., 2019] and evaluate the effects of window size on subjective impressions [Moscoso et al., 2021]. A fisheye rendering with 180-degree FOV shows the full scope of the interior scene perceived by the human eyes. Using a digital office model, Jakubiec and Reinhart rendered 360-degree panoramic scenes to explore a single occupant's flexible view directions through point-in-time and annual luminance simulations [Jakubiec and Reinhart, 2012]. Later, Hashemloo et al. rendered indoor luminance images and used the overlit areas to design the shading strategies at different floor orientations, only focusing on the point-in-time simulation results [Hashemloo et al., 2016].

Too much light entering through windows can cause excessive contrast and glare problems. As glare is view-dependent, it can be difficult to predict glare experienced by multiple users at different locations. The existing challenges in luminance analysis include:

- multiple view directions due to different computer screens
- multiple view heights due to adjustable standing desks
- occupants' different luminance thresholds due to individual preferences

These challenges necessitate a computational method that integrates multiple design parameters and synthesizes the results from different simulation results.

### 2.3. Objectives

This study considers various view directions, positions, and luminance thresholds in a typical open workplace. The workflow illustrates using multiple inputs (such as view direction, height and luminance thresholds) to develop facade patterns. Specifically, the objectives of this study are:

- To develop a geometry-based method to project a 2D fisheye image into the 3D space.
- To perform batch luminance simulation for point-in-time and annual climate-based modes.
- To use luminance threshold to outline the facade areas that emit excessive luminance from different view positions and directions.

## 3. METHODOLOGY

### 3.1. Building Model

A 182.0 m$^2$ office room (Fig. 2) was selected as the base case, with dimensions of 10.0m (width) by 18.2m (length) by 3.7m (height). The model is provided by Climate Studio software [Solemma LLC, 2021] built in Rhino [Robert McNeel & Associates, 2021] and Grasshopper [Grasshopper - algorithmic modeling for Rhino, 2022]. The original model was adjusted to include two computer monitors at each desk, and the desks were set to vary in height. The South elevation was selected for luminance mapping.

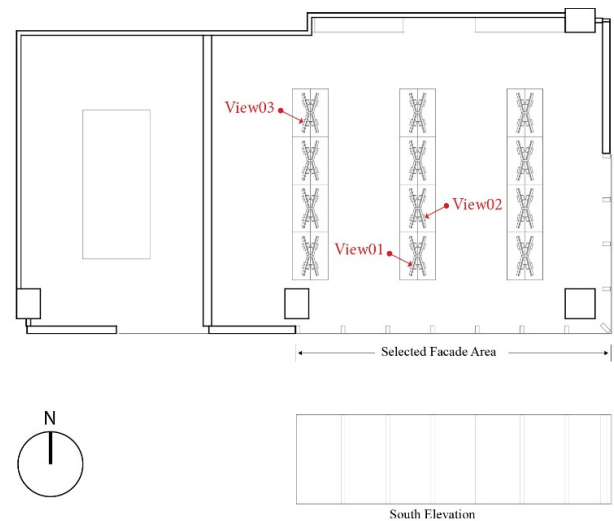

Fig. 2. Building Model Floor Plan and South Elevation

Three view positions were selected in the model. To test different view directions in the workplace, each view position is set to focus on a different computer monitor. View01 is the closest to the South facade with a desk height of 0.7m, focusing on the right computer monitor. View02 has a desk height of 0.7m and is set to focus on the left computer monitor. View03 has a desk height of 1.0 m and focuses on the right computer monitor. Figure 2 describes the three occupants in 3 different office locations.

### 3.2. Simulation Settings

The simulation analyses were conducted in Climate Studio [Solemma LLC, 2021], built on Radiance which is a physics-based rendering system built on a light-backwards ray-tracing approach [Ward, 1994]. For architectural lighting simulation, Radiance has been validated for estimating the interior light levels under different sky conditions [Mardaljevic, 1995]. Lighting simulations used the CIE clear sky condition and local weather file (USA_PA_Pittsburgh-Allegheny.County.AP.725205_TMY3.epw). The simulation parameters are listed in TABLE I. Window blinds and electrical lighting devices were not included in the model. Detailed material properties are listed in the Appendix (TABLE II).

TABLE I. SIMULATION PARAMETERS AND ASSIGNED VALUES

| ambient bounces (ab) | ray weight (lw) | samples per pixel | image dimensions |
|---|---|---|---|
| 8 | 0.01 | 100 | 400 (Height) 400 (Width) |

As shown in Fig. 3, a vector is established by two points; a point on the center of computer screen and a point at eye level. This vector describes the view direction when the occupant sits in front of the desk and looks at one of the computer monitors. In this study, FOV is fixed at 180 (180-degree in both horizontal and vertical directions).

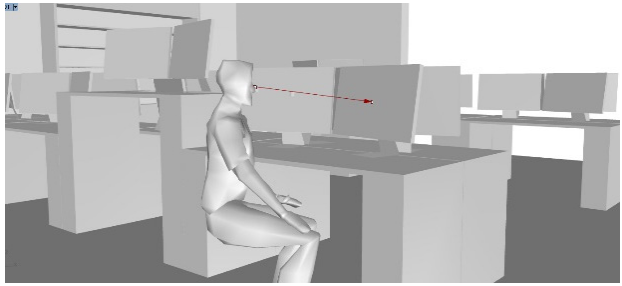

Fig. 3.  View Setting (View01)

Previous studies on visual comfort used different luminance thresholds. One simulation study used 2000 cd/m² as the absolute luminance threshold in the office environment [Van Den Wymelenberg et al., 2010]. In another field study, a 3200 cd/m² luminance value was observed as the upper limit when users adjusted the blinds to achieve visual comfort [Sutter et al., 2006]. A recent simulation study of luminance projection in an office space defined 3000 cd/m² as the threshold for discomfort glare [Hashemloo et al., 2016].

This study used 3000 cd/m² as the upper threshold for generating the initial rendered images. In these greyscale fisheye images, a pixel value of 0 corresponds to 0 cd/m² and a pixel value of 255 corresponds to 3000 cd/m². When testing the impacts of different thresholds on facade design, the target luminance value will be approximated by filtering pixel values proportionally. As previously mentioned, several thresholds have been used in the literature, but the most common one is 2000 cd/m² [Pierson et al., 2018]. Therefore, in this study, pixels having a luminance value higher than 2000 cd/m² are treated as the potential glare source.

### 3.3. Image Processing

A digital image obtained from a luminance simulation typically contains noise. White noise will cause errors when filtering high pixel values. Bilateral Filtering [Tomasi and Manduchi, 1998] was developed for smoothing images based on the nonlinear combination of surrounding pixel values. The method can remove the noise in the image and preserve the sharp edges. In this study, the Bilateral Filter function (sigmaColor = 75, sigmaSpace = 75, diameter of each pixel neighborhood = 15) from the OpenCV library [Intel Corporation, 2021] was used to smooth the original images, and the effect of Bilateral Filtering is presented in Fig. 4.

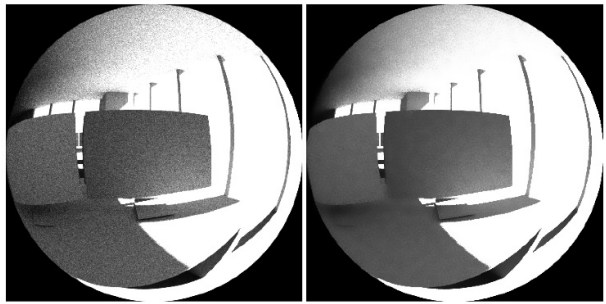

Fig. 4.  Left: Original Rendered Image, Right: Image Smoothed by Bilateral Filtering

The existing building standards and metrics use time frequency to highlight the most problematic areas or times when glare occurs. To highlight the facade areas that allow too much light (over 2000 cd/m²) over an extended period, a frequency test was conducted. When computational approaches simulated the annual daylight performance in 8760 hours, few of the previous studies just selected daylight hours [Inanici, 2021]. Based on batch simulation results, all rendered images are binarized by the target luminance threshold (2000 cd/m²). The sum of each pixel value from the binarized images was divided by the number of superimposed images (10 images) to ensure the value of each pixel was within the range of 0-255. In this case, the pixel value can be regarded as time frequency (Fig. 5). Any pixel value that is over the luminance threshold for more than the listed percentile is colored in red. As can be seen in Fig 5, when the time frequency is set to 5%, most of the pixels are selected. With the increase of time frequency, fewer pixels are colored. The result of the 95 percentile only highlighted a part of the glazing and ground area, while the 50 percentile identified some high luminance values from both ceiling and floor. This study chose the 50 percentile to examine the target luminance (2000 cd/m²).

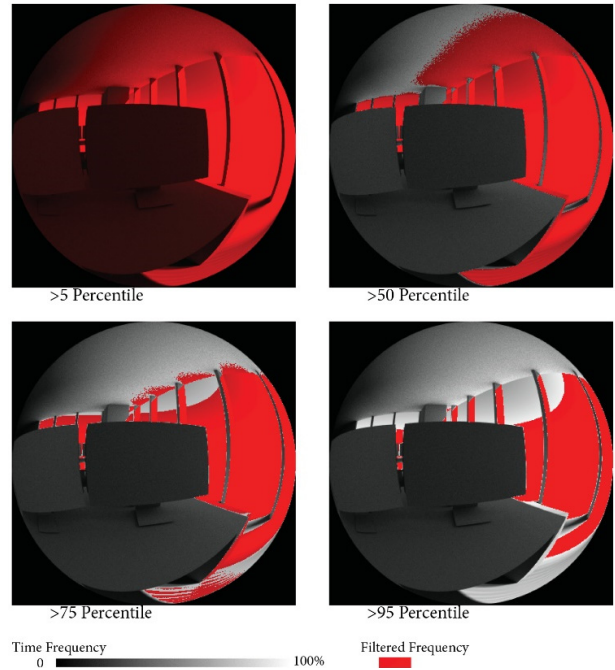

Fig. 5.  Alternatives for Time Frequency

### 3.4. Luminance Mapping Workflow

As shown in Fig. 1 (a), this study first rendered a fisheye image from an occupant's position. Then, a target luminance threshold (2000 cd/m²) was used to divide all the pixels into white (above threshold) and black (under threshold). Image pixels were converted into 3D spherical coordinates. As illustrated in Fig.6, each point (p) on the hemisphere can be expressed as:

$$p = \begin{bmatrix} x \\ y \\ z \end{bmatrix} = \begin{bmatrix} r\sin\varphi\cos\theta \\ r\sin\varphi\sin\theta \\ r\cos\varphi \end{bmatrix} \quad (1)$$

Where $r$ is the sphere radius, $\theta$ is the polar angle, and $\varphi$ is the azimuthal angle.

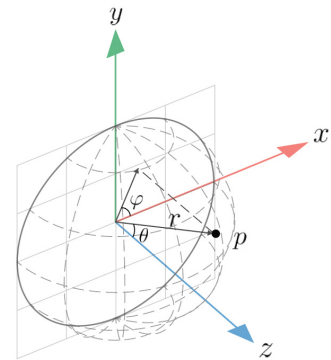

Fig. 6.  Point Coordinates on Hemisphere

After the view vector is defined and pixels are projected into a 3D space, the rendered fisheye image (Fig. 1 (a)) is mapped onto the hemispherical surface (Fig. 1 (b)). On the hemisphere, target pixels are projected into 3D space (Fig. 1 (c)), and the projected geometries will outline the facade areas that emit high luminance values in the given view position (Fig. 1 (d)). Higher image resolution results in longer computation time. To increase computation efficiency, when converting image coordinates, image resolution is reduced from 400 (Height) by 400 (Width) to 80 (Height) by 80 (Width); this resolution maintained the edges of pixel areas and achieved efficient computation time (5.1 seconds) for a single image. Fig.7 shows four image resolutions and the corresponding computation time on Intel(R) Core(TM) i7-8650U CPU.

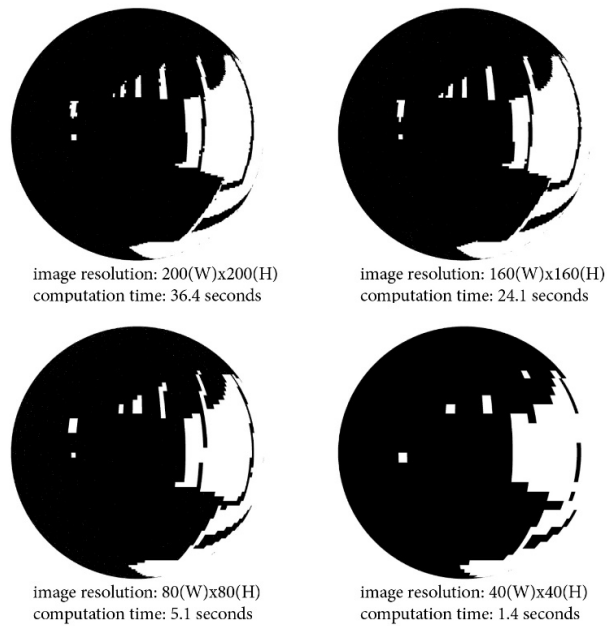

Fig. 7. Image Resolution and Computation Time

## 4. RESULT

### 4.1. Single View in Point-in-Time Simulation

Current climate data and simulation software do not support small time intervals (such as 1 minute) in daylight calculations. To address this shortcoming, the point-in-time simulation was set to run from 1:00-2:00 PM on March 21st. The developed simulation algorithm divides the one-hour simulation into 10 intervals. Under this setting, the scene was simulated every 6 minutes.

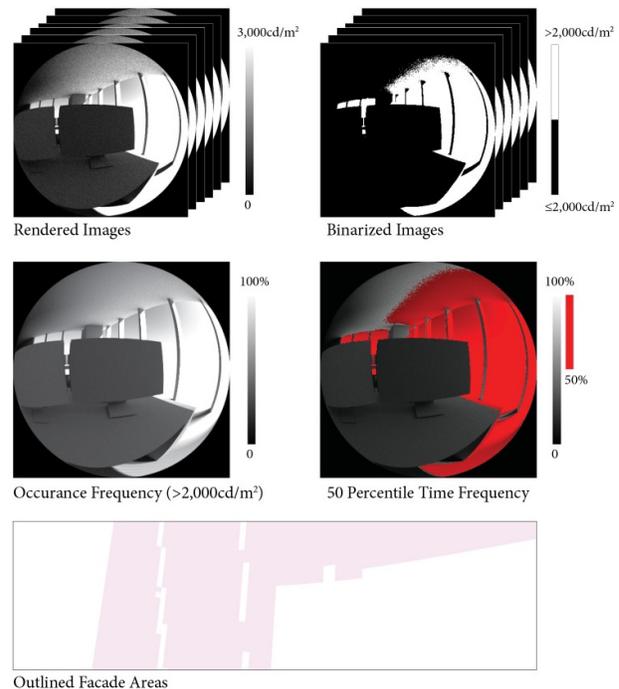

Fig. 8. Image Processing Workflow

As shown in Fig. 8, all rendered images were binarized using 2000 cd/m$^2$ as the benchmark. The luminance values above 2000 cd/m$^2$ were colored in white, and any pixels below 2000 cd/m$^2$ were colored in black. The result highlighted the areas where luminance above 2000cd/m$^2$ occurred over 50% percent of the selected daytime hours. The red regions reflect the target areas and the projected red pixels outline the corresponding facade areas.

### 4.2. Single View in Different Months

Sun's position and direction change over time, causing the indoor luminance distribution to vary in different months. This section focuses on the results of luminance simulations for each month. The simulation period was from 9:00 AM to 6:00 PM at every one hour. The total number of simulated images was 10 (hours) multiplied by the number of days in each month. Following the image processing workflow (Fig. 8), Fig. 9 illustrates the results from View01 in 12 months using 2000cd/m$^2$ as the luminance threshold and highlights the pixels that receive over 2000 cd/m$^2$ for more than 50% of the time.

The results indicate that larger overlit areas colored in red and outlined on the glazing surface between March and September, compared to the other months. Without the surrounding buildings modeled in this study, the ground surface reflects too much light between February and October. During the winter, the desk surface receives excessive light due to the lower sun's altitude. The floor also shows overlit areas in February, March, April, September, and October. Fig. 10 illustrates the projected pixels on the facade surface in different months. The results between February and October show the glazing area in the middle to the left of the room consistently allows extra light to enter the workplace, while the overlit areas are smaller in January, November, and December.

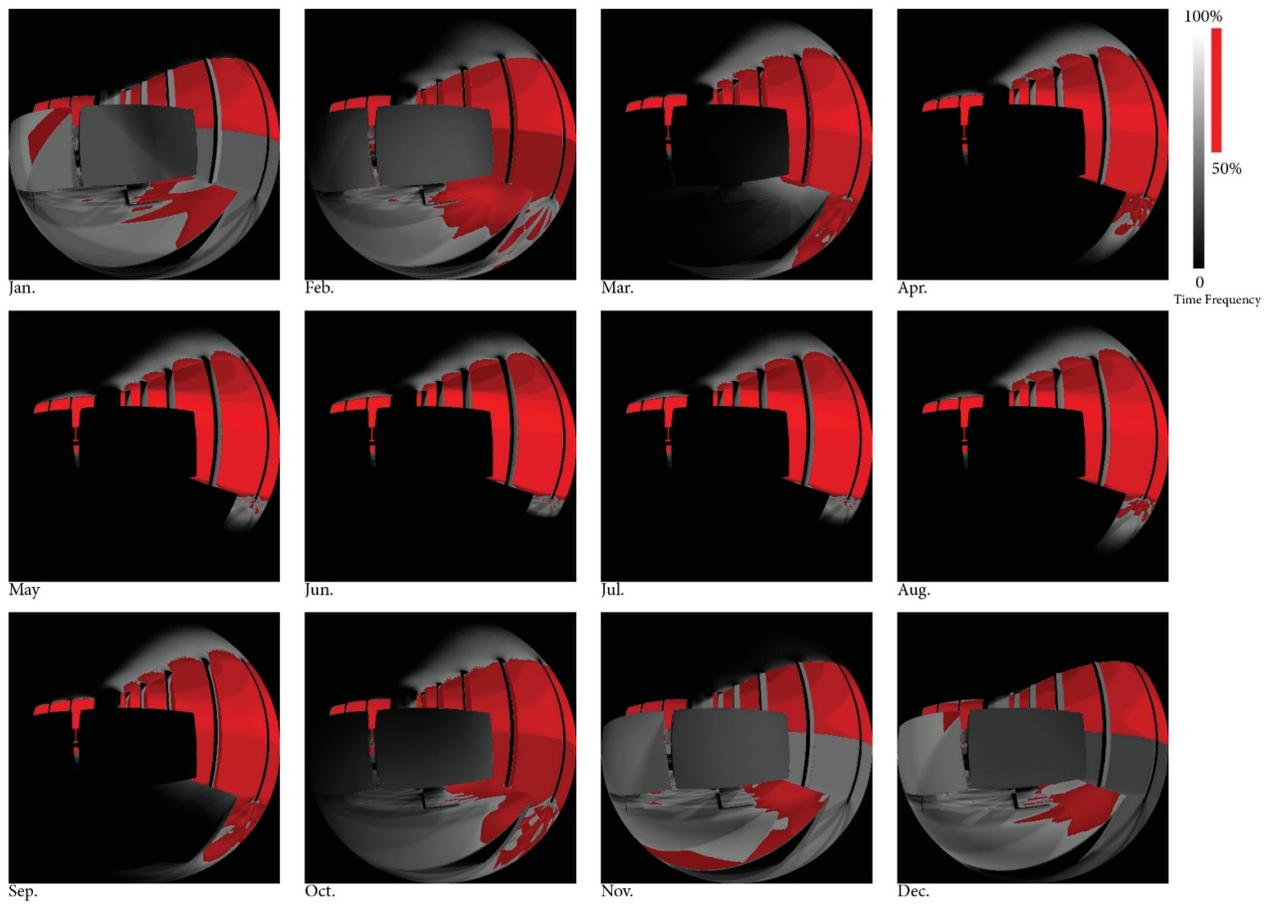

Fig. 9. Single View in 12 Months

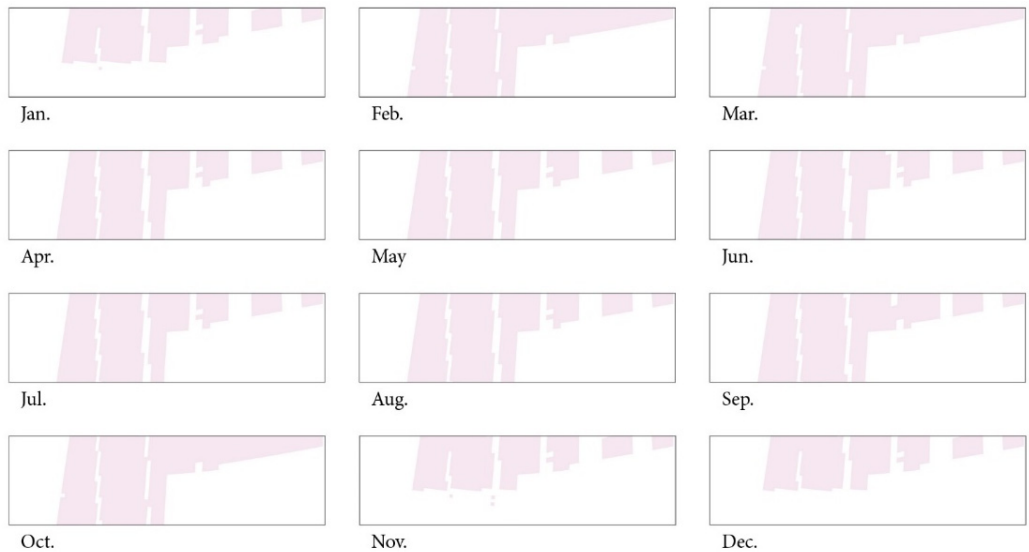

Fig. 10. Outlined Facade Areas in 12 Months

### 4.3. Multiple Views in Different Months

View01, View02, and View03 used 2000cd/m$^2$ and 50% as target luminance and time frequency, respectively. The luminance simulation ran for 31 days in March from 9:00 AM to 6:00 PM at 10 intervals. In total, 310 rendered images were collected from each view position. Fig. 11 (b) shows the results from three views (Fig.11 (a)). In Fig.11 (c), View03 shows the smallest overlit areas among the three view positions.

Due to the different view directions, although view01 and view02 outlined different areas on the facade, the middle area of the facade shows consistent overlit areas. Fig. 11 (d) shows the facade surface when three projected facade patterns overlapped. The three magenta colors illustrate the overlit areas by the number of views (one, two, and three), and the overlapped regions represent the upper middle areas that regularly allow too much light to enter the workplace, potentially causing glare.

To analyze facade patterns in different seasons, March, June, and December were selected for luminance simulation. The results for June and December are shown in Fig. 12, following the same settings as Fig. 11. Similar to March, the upper middle glazing area emits excessive light from the three views in June. For December, the coverage of outlined areas was smaller than in March and June, and the lower part of the facade area was excluded.

## 5. Conclusion

### 5.1. Occupant-Centric Luminance Analysis

This study proposes an occupant-centric view analysis approach that uses fisheye images to project pixels with high luminance from the occupant's perspective to the facade to improve daylight performance. The workflow provides a technique to transform two-dimensional fisheye images into three-dimensional surfaces. It also allows for analyzing a large number of rendered scenes under different parameters and provides an image processing workflow that highlights the areas of the glazing that emit too much direct sunlight into the workplace.

Such an approach is effective in studying instantaneous luminance variations over a long period of time. While the results of this study only focused on pixels that received above 2000 cd/m$^2$ over 50% of the daytime hours, the workflow could easily implement other thresholds for further analysis.

### 5.2. Applications in Design

Although this study only tested three view positions in an open office plan, future studies could include more occupants, view settings, and simulation parameters. By establishing a view vector for different occupants, the computational method can be applied to other building typologies, including hospitals, schools, and residences. The workflow could be used to study indoor scenarios in different climate zones, weather conditions, and time periods. The 3D modeling method provides a new approach to guide a high-performance facade design, and the outlined facade areas can be utilized to create novel facade patterns, including internal and external shading devices.

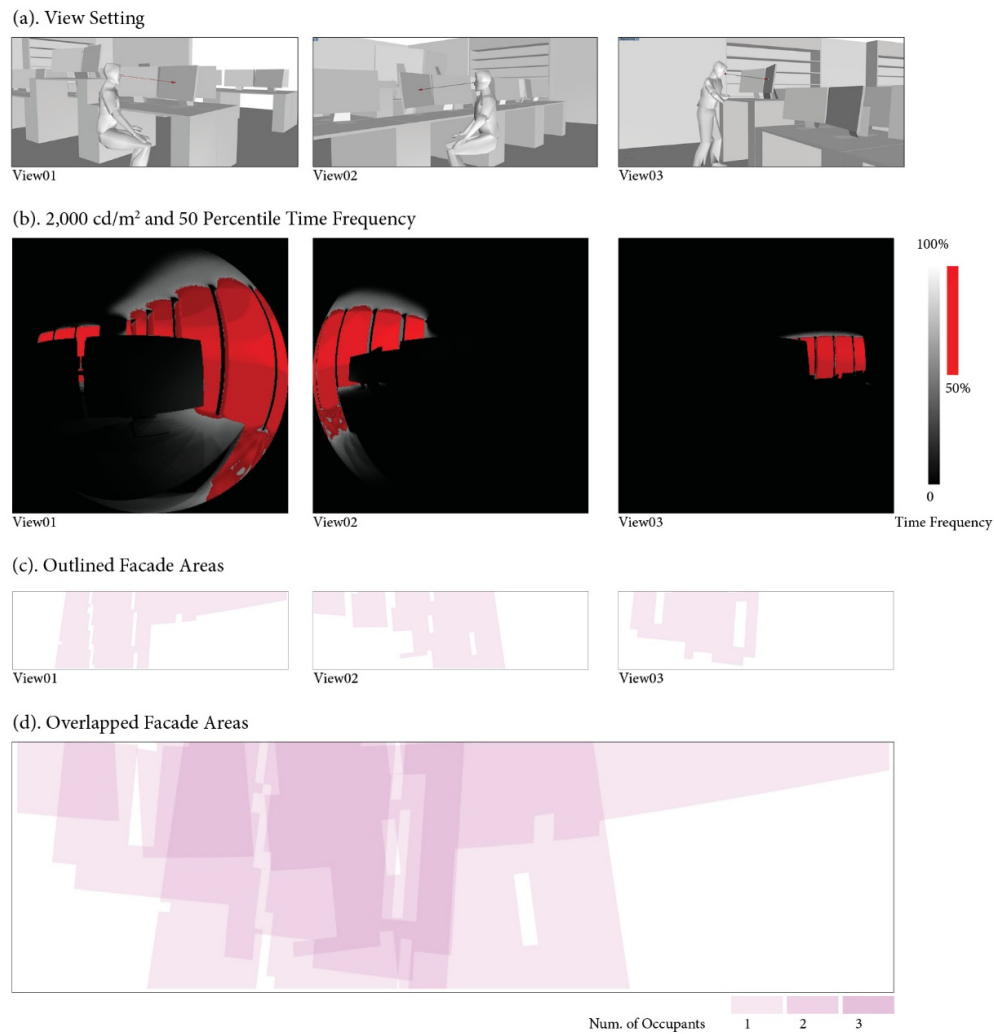

Fig. 11. Multiple Views in Mar.

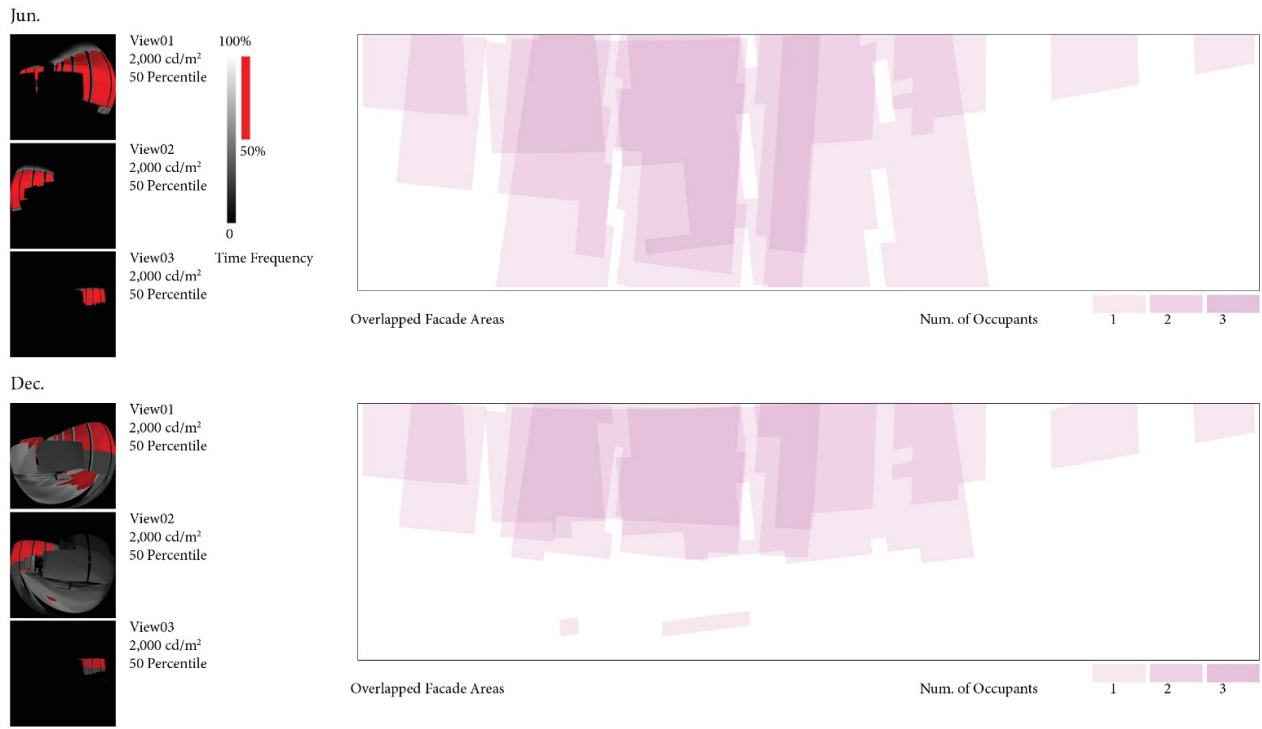

Fig. 12. Multiple Views in Jun. and Dec.

## PART 2: IMPLEMENTATION [I/O SECTION]

### 6. IMAGE PROCESSING ALGORITHM

The image processing approach is written in Python with open-source libraries. The inputs are a large number of rendered images (Fig. 13). Fig. 14 and Fig. 15 show the intermediate results when running the code.

#### 6.1. input

```
# inport python packages
import glob, os
import matplotlib.pyplot as plt
from PIL import Image
import cv2
import numpy as np
import os.path
```

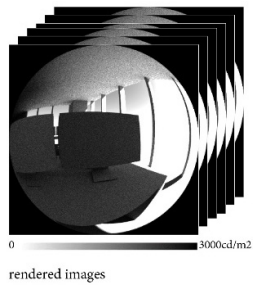

Fig. 13. simulated luminance images with noises

```
#define threshold
threshold = 2000
luminance_cap = 3000
ratio = threshold/luminance_cap
#define image processing function
def img_process(orig_img):
    #read img
    img = cv2.imread(orig_img)
    height, width = img.shape[0], img.shape[1]
    img = cv2.bilateralFilter(img, 15, 75, 75)
    new_img = np.zeros((height, width, 3), np.uint8)
    # Binarization
    for i in range(height):
        for j in range(width):
            for k in range(3):
                if img[i, j][k] < ratio*255:
                    gray = 0
                else:
                    gray = 255
                new_img [i, j][k] = np.uint8(gray)
    return new_img

#define directory and folder path
Directory = "path for the directory"
```

```
folder0 = "path for the noisy images"
folder1 = "path for the denoised images"

#batch image processing
for filename in os.listdir(Directory+folder0):
    img = Directory " + folder0 + filename
    print(Directory + filename)
    post_img = img_process(img)
    cv2.imwrite(Directory+folder1+filename + ".png", post_img)
```

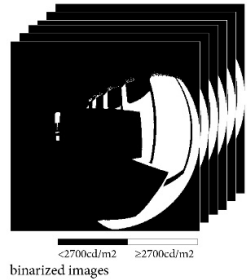

Fig. 14. Images after Denoising and Binarization

```
#create empty matrix to store matrix information
img_compile = np.zeros((400, 400))
num = len(os.listdir(Directory+folder1))
#read and combine all images in the folder
for filename in os.listdir(Directory+folder1):
    #for i in range(num):
    img_dir = Directory + folder1 + filename
    img_r = cv2.imread(img_dir, 0)
    img_compile = img_compile+img_r
 #avarage the pixel values
img_compile = img_compile/num
```

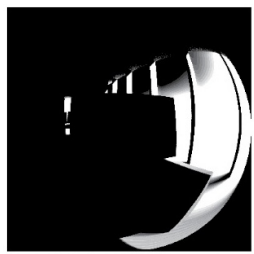

Fig. 15. Combined Images

```
#filter the occurrence frequency
percentile = 0.50
img=cv2.imread(path_of_combined_image, 0)
shape = img.shape
result = np.zeros(shape)
for x in range(0, shape[0]):
    for y in range(0, shape[1]):
        if img[x, y] >= 255*percentile:
            result[x, y] = 255
        else:
            result[x, y] = 0
channel = np.ones((400, 400))
red_mask = cv2.merge([channel, channel, result])
binary_mask = result
img_seg = img - result
img_3channels = cv2.merge([img_seg, img_seg, img_seg])
final_img = img_3channels + red_mask
cv2.imwrite path_for_image + "binary.png", binary_mask)
cv2.imwrite(path_for_image + "final.png", final_img)
```

*6.2. Output*

Fig.16 shows the output images after running the code.

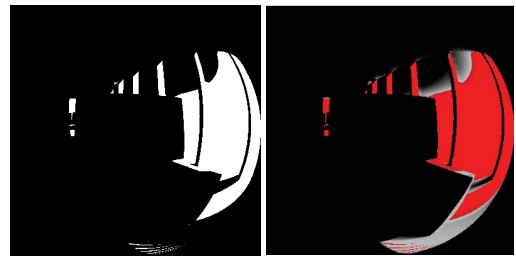

Fig. 16. Left: Binary Image, Right: Colored Final Image

**7. LUMINANCE MAPPING ALGORITHM**

The workflow is built on Grasshopper platform in Rhino. The inputs include two points and one image. Two points are the view start point and view end point for rendering the fisheye image (Fig. 17). As shown in Fig. 18, the input image is vertically flipped from the original binary image.

*7.1. Input*

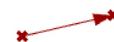

Fig. 17. Two Points for View Vector

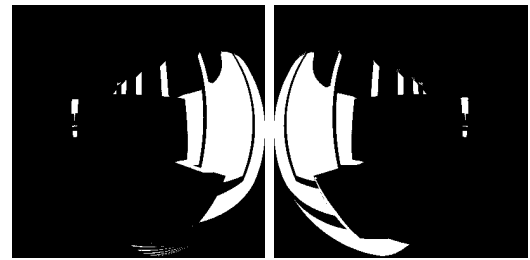

Fig. 18. Left: Binary Image, Right: Flipped Binary Image

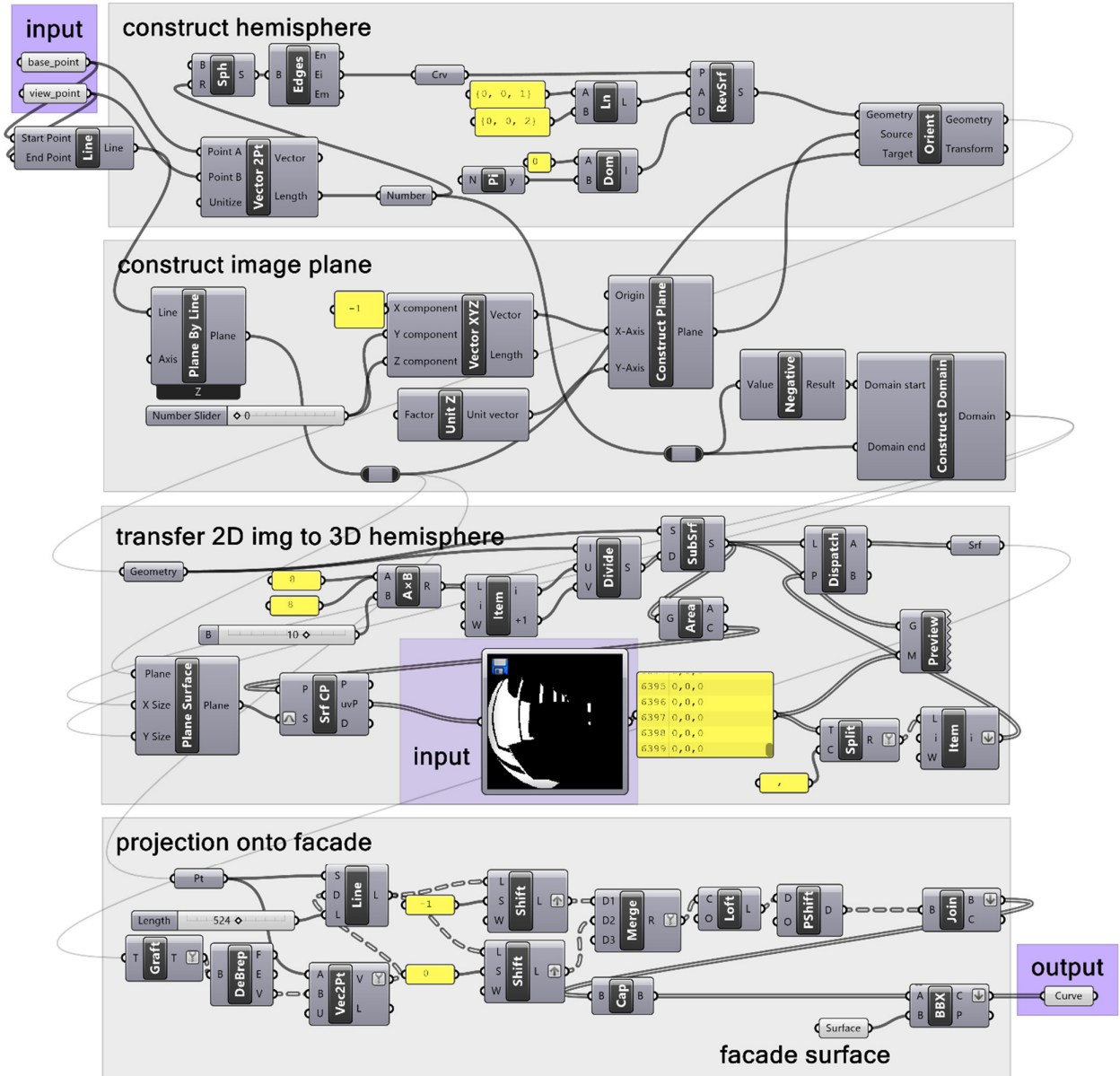

Fig. 19. Grasshopper Workflow

### 7.2. Output

The output from the Grasshopper workflow (Fig. 19) illustrates the outlined facade areas (Fig. 20).

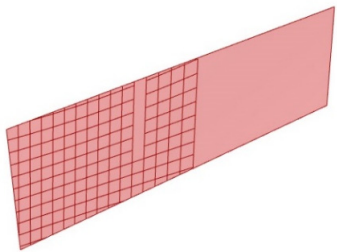

Fig. 20. Outlined Facade Areas

## 8. APPENDIX

TABLE II.    OBJECTS AND MATERIAL PROPERTY

| Opaque Material | Material Name | Red Reflectance | Green Reflectance | Blue Reflectance | Specular | Roughness |
|---|---|---|---|---|---|---|
| Mullion | Wall LM83 | 0.5 | 0.5 | 0.5 | 0 | 0 |
| Ceiling | Ceiling LM83 | 0.7 | 0.7 | 0.7 | 0 | 0 |
| Floor | Floor LM83 | 0.2 | 0.2 | 0.2 | 0 | 0 |
| Wall | Wall LM83 | 0.5 | 0.5 | 0.5 | 0 | 0 |
| Furniture | Furniture LM83 | 0.5 | 0.5 | 0.5 | 0 | 0 |
| **Glazed material** | **Material Name** | **Tavis** | **Rvis.front** | **Rvis.back** | **U-value** | **SHGC** |
| Glazing | Clear-Clear, Double Layer | 0.774 | 0.149 | 0.150 | 2.69W/m2K | 0.703 |

*LM83: IES standard LM-83-12 [ Illuminating Engineering Society of North America (IESNA),2012]